\documentclass{article}

 \usepackage[preprint]{neurips_2025}

\usepackage[utf8]{inputenc} % allow utf-8 input
\usepackage[T1]{fontenc}    % use 8-bit T1 fonts
\usepackage{hyperref}       % hyperlinks
\usepackage{url}            % simple URL typesetting
\usepackage{booktabs}       % professional-quality tables
\usepackage{amsfonts}       % blackboard math symbols
\usepackage{nicefrac}       % compact symbols for 1/2, etc.
\usepackage{microtype}      % microtypography
\usepackage{xcolor}         % colors

\usepackage{algorithm}
\usepackage{algorithmic}
\usepackage{multirow}
\usepackage{colortbl}
\usepackage{amsfonts} % For \mathbb
\usepackage{amsmath}
\usepackage{amssymb}
\usepackage{graphicx}

\title{GPG: Generalized Policy Gradient Theorem for Transformer-based Policies}

\author{%
  Hangyu Mao  \\
  \texttt{hy.mao@pku.edu.cn} \\
  \And
  Guangting Dong \\
  \texttt{dongguanting@ruc.edu.cn} \\
  \And
  Zhicheng Dou \\
  \texttt{dou@ruc.edu.cn} \\
}

\begin{document}

\maketitle

\begin{abstract}
We present the \textbf{Generalized Policy Gradient (GPG) Theorem}, specifically designed for Transformer-based policies. Notably, we demonstrate that both standard Policy Gradient Theorem and GRPO emerge as special cases within our GPG framework. Furthermore, we explore its practical applications in training Large Language Models (LLMs), offering new insights into efficient policy optimization.
\end{abstract}

\section{Introduction}
Proximal Policy Optimization (PPO) \cite{schulman2017proximal} and Group Relative Policy Optimization (GRPO) \cite{shao2024deepseekmath} rank among the most widely adopted policy gradient algorithms for training Large Language Models (LLMs), which predominantly employ the Transformer architecture \cite{vaswani2017attention}. However, these algorithms were initially developed for general reinforcement learning (RL) policies \cite{sutton1998reinforcement} rather than being specifically optimized for Transformer-based policies.

Given the pivotal role of LLMs in AI research, we address a critical question: \textbf{Are there policy gradient methods inherently better suited for training Transformer-based LLM policies?} We hypothesize that specialized algorithms could surpass generic policy gradient approaches in both theoretical alignment and empirical performance.

Our primary contribution is a Generalized Policy Gradient (GPG) Theorem tailored for Transformer-based policies. We prove that both the standard Policy Gradient Theorem and GRPO emerge as concrete implementations derived from our GPG framework. Additionally, we investigate practical applications of the GPG Theorem for LLM training, offering insights into its potential advantages over existing methods.

\section{Preliminary}
We consider the standard RL framework \cite{sutton1998reinforcement}, in which a learning agent interacts with an environment. The state, action, and reward at each timestep $t$ are denoted by $s_t$, $a_t$, and $r_t$, respectively. The environment is characterized by the state transition probabilities $P(s_{t+1}|s_t, a_t)$ and the reward function $r_t = R(s_t,a_t)$. The agent's decision making procedure at each timestep is characterized by a stochastic policy $\pi_{\theta}(s_t|a_t) = P(a_t|s_t; \theta) \in [0, 1]$, with the objective function to maximize the long-term cumulative reward $R(\tau) = \sum_{t=1}^{H} r_t$ , where $\tau = \langle s_1, a_1, r_1, ..., s_H, a_H, r_H \rangle$ is the decision trajectory and $H$ is the horizon.

\subsection{Policy Gradient Theorem}
Define the parameterized policy $\pi_{\theta}(a_t|s_t)$ and its objective function $J(\theta)=\mathbb{E}_{\tau \sim \pi_\theta}[R(\tau)]$. We use gradient acsend to find the optimal $\theta^{*}$ that can maximize the objective:
\begin{equation}
    \theta \leftarrow \theta + \alpha * \nabla_{\theta} J(\theta)
\end{equation}

The Policy Gradient Theorem \cite{sutton1999policy} says that for any differentiable policy and any objective function, the gradient of the parameterized policy is as follows:
\begin{equation}
    \nabla_{\theta} J(\theta) 
    = \mathbb{E}_{\tau \sim \pi_\theta}\{ \sum_{t=1}^{H}[\nabla_{\theta}\log\pi_{\theta}(a_t|s_t) R(\tau)] \} \label{equ:PG1}
\end{equation}

A general form of the Policy Gradient Theorem is:
\begin{equation}
\begin{aligned}
    \nabla_{\theta} J(\theta) 
    = & \mathbb{E}_{\tau \sim \pi_\theta}\{ \sum_{t=1}^{H}[\nabla_{\theta}\log\pi_{\theta}(a_t|s_t) \Phi_t ] \} \label{equ:PG2}
\end{aligned}
\end{equation}
where $\Phi_t$ may be one of the following:
\begin{itemize}
    \item $\sum_{t=1}^{H} r_t$: total reward of the trajectory.
    \item $\sum_{t'=t}^{H} r_{t'}$: reward following action $a_t$.
    \item $\sum_{t'=t}^{H} r_{t'} - b(s_t)$: baselined version of previous formula.
    \item $Q^{\pi_\theta}(s_t, a_t)$: state-action value function.
    \item $A^{\pi_\theta}(s_t, a_t)$: advantage function.
    \item $r_t + V^{\pi_\theta}(s_{t+1}) - V^{\pi_\theta}(s_t)$: Temporal-Difference residual.
    \item $A^{\text{GAE}(\gamma,\lambda)}(s_t, a_t)$: the generalized advantage estimator (GAE) for the advantage function.
\end{itemize}

In practice, the advantage function $A^{\pi_\theta}(s_t, a_t)$ is a common choice because it can achieve better bias-variance trade-off \cite{schulman2015high}.

\subsection{TRPO, PPO and GRPO}
TRPO \cite{schulman2015trust}, PPO \cite{schulman2017proximal} and GRPO \cite{shao2024deepseekmath} are the special implementations of the Policy Gradient Theorem. PPO and its predecessor TRPO  optimize policies (i.e., getting the new policies) with guaranteed monotonic improvement by considering the trust region of old policies. In practice, this is implemented with the off-policy importance sampling strategy (i.e., using old policies $\pi_{\theta_{\text{old}}}$ to sample trajectories to estimate the gradient of new policies $\pi_{\theta}$):
\begin{equation}
\begin{aligned}
    \nabla_{\theta} J(\theta) = & \mathbb{E}_{\tau \sim \pi_\theta}\{ \sum_{t=1}^{H}[\nabla_{\theta}\log\pi_{\theta}(a_t|s_t) A^{\pi_\theta}(s_t, a_t) ] \} \\
    = & \mathbb{E}_{\tau \sim \pi_{\theta_{\text{old}}}}\{ \sum_{t=1}^{H}[\frac{\pi_{\theta}(a_t|s_t) \rho_{\theta}(s_t)}{\pi_{\theta_{\text{old}}}(a_t|s_t) \rho_{\theta_{\text{old}}}(s_t)} \nabla_{\theta}\log\pi_{\theta}(a_t|s_t) A^{\pi_\theta}_t ] \} \\
    = & \mathbb{E}_{\tau \sim \pi_{\theta_{\text{old}}}}\{  \sum_{t=1}^{H}[\frac{\nabla_{\theta}\pi_{\theta}(a_t|s_t) \rho_{\theta}(s_t)}{\pi_{\theta_{\text{old}}}(a_t|s_t) \rho_{\theta_{\text{old}}}(s_t) } A^{\pi_\theta}(s_t, a_t) ] \}  \\
    \approx & \mathbb{E}_{\tau \sim \pi_{\theta_{\text{old}}}}\{  \sum_{t=1}^{H}[\frac{\nabla_{\theta}\pi_{\theta}(a_t|s_t)}{\pi_{\theta_{\text{old}}}(a_t|s_t) } A^{\pi_{\theta_{\text{old}}}}(s_t, a_t) ] \}
\end{aligned}
\end{equation}
where $\rho_{\theta}(s_t)$ is the station state distribution under policy $\pi_\theta$. Therefore, the (unclipped) off-policy ``surrogate'' objective can be represented as:
\begin{equation}
\begin{aligned}
    J(\theta) = & \mathbb{E}_{(s_t,a_t) \sim \pi_{\theta_{\text{old}}}}[\frac{\pi_{\theta}(a_t|s_t)}{\pi_{\theta_{\text{old}}}(a_t|s_t)} A^{\pi_{\theta_{\text{old}}}}(s_t, a_t) ] \\
    = & \mathbb{E}_{(s_t,a_t) \sim \pi_{\theta_{\text{old}}}}[r_t(\theta) A^{\pi_{\theta_{\text{old}}}}(s_t, a_t) ]
\end{aligned}
\end{equation}

However, ``without a constraint, maximization of the above $J(\theta)$ would lead to an excessively large policy update'' \cite{schulman2017proximal}, hence, PPO also applies the region clip strategy to penalize changes to the policy that move the ratio $r_t(\theta)$ away from 1. So the formal (clipped) objective of PPO is:
\begin{equation}
\begin{aligned}
    J(\theta) 
    = \mathbb{E}_{(s_t,a_t) \sim \pi_{\theta_{\text{old}}}}[ \min( r_t(\theta)  A^{\pi_{\theta_{\text{old}}}}(s_t, a_t), \text{clip}(r_t(\theta), 1-\epsilon, 1+\epsilon) A^{\pi_{\theta_{\text{old}}}}(s_t, a_t) ) ] \label{equ:PPO}
\end{aligned}
\end{equation}
where $\epsilon$ is a small value such as 0.1.

\begin{figure*}[!t]
    \centering
    \includegraphics[width=1.0\linewidth]{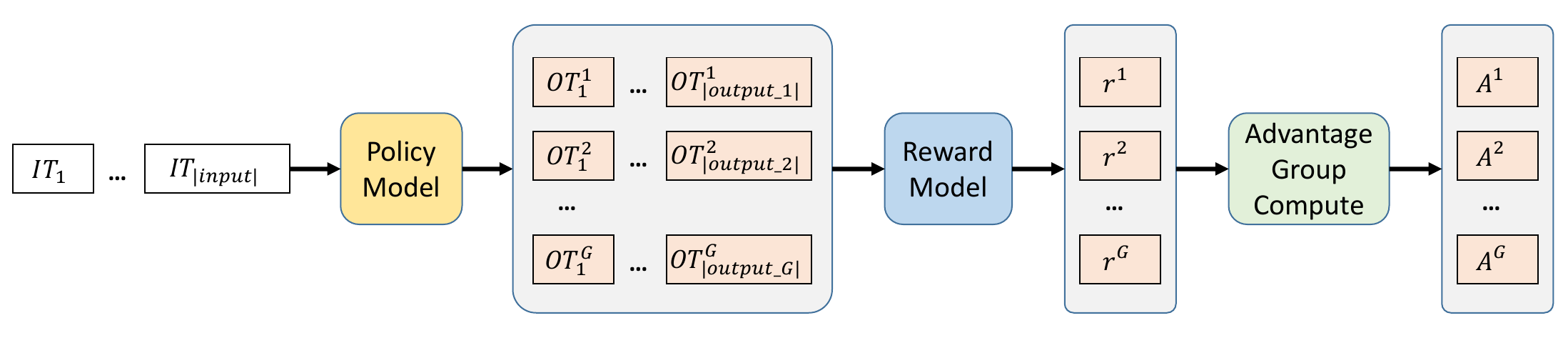}
    \caption{The computing process of GRPO.}
    \label{fig:GRPO}
\end{figure*}

GRPO shares the same objective as PPO, but uses a group of $G$ output trajectories to compute the advantage $A^{\pi_{\theta_{\text{old}}}}(s,a)$ as shown in Figure \ref{fig:GRPO}. This is especially useful for scenarios (e.g., math and coding) where the partial trajectory is not verifiable, but the whole trajectories can be evaluated with verified rewards.

\subsection{Chain Rule}
The chain rule of probability theory states that for any joint probability distribution $P(x_1, x_2, ..., x_n)$, the following decomposition holds:
\begin{equation}
\begin{aligned}
    P(x_1, x_2, ..., x_n) = & P(x_1) \times P(x_2|x_1) \times ... \times P(x_n|x_1, x_2, ..., x_{n-1})
\end{aligned}
\end{equation}

Similarly, for any conditional joint probability distribution $P(x_1, x_2, ..., x_n|y_1, y_2, ..., y_m)$, the chain rule yields the following decomposition:
\begin{equation}
\begin{aligned}
    P(x_1, x_2, ..., x_n|y_1, y_2, ..., y_m) = & P(x_1|y_1, y_2, ..., y_m) \times \\ 
    & P(x_2|y_1, y_2, ..., y_m, x_1) \times \\
    & P(x_3|y_1, y_2, ..., y_m, x_1, x_2) \times \\
    & ... \\
    & P(x_n|y_1, y_2, ..., y_m, x_1, x_2, ..., x_{n-1})
\end{aligned}
\end{equation}

\section{Generalized Policy Gradient Theorem}
\begin{figure}[!th]
    \centering
    \includegraphics[width=0.5\linewidth]{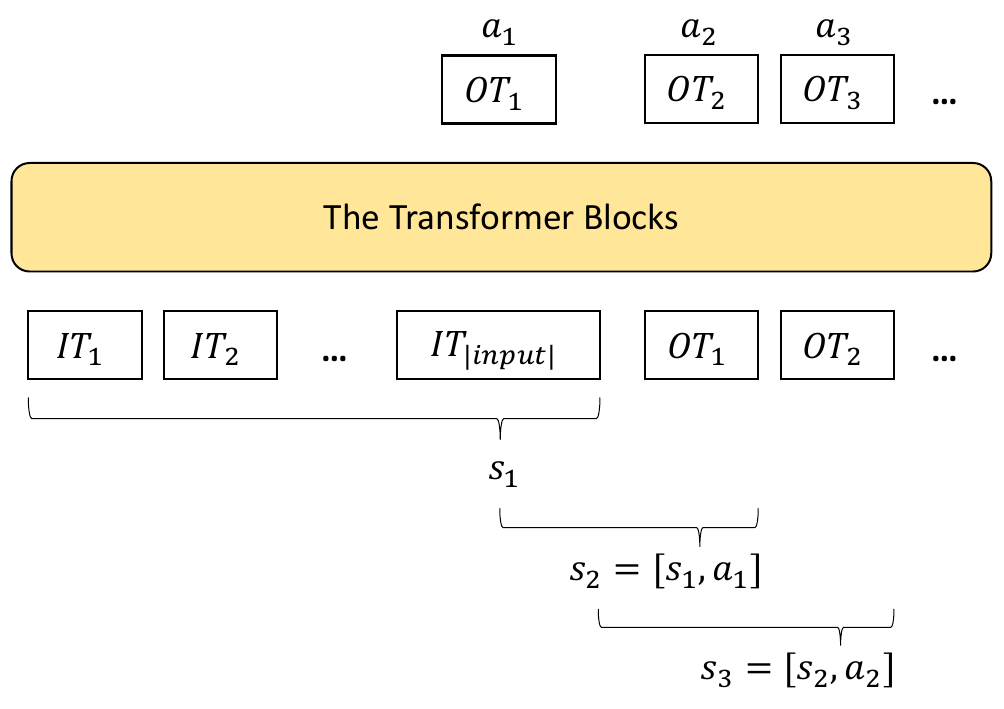}
    \caption{The illustration of Transformer-based polices.}
    \label{fig:TransformerPolicy}
\end{figure}
\subsection{Transformer-based Policy}
The Transformer-based policy $\pi_{\theta}(a_t|s_t)$, as illustrated in Figure \ref{fig:TransformerPolicy}, can be decomposed via the chain rule as follows:
\begin{equation}
\begin{aligned}
    & \pi_{\theta}(OT_1 \;\;|\;\; IT_1, IT_2, ..., IT_{|input|}) \times \\
    & \pi_{\theta}(OT_2 \;\;|\;\; IT_1, IT_2, ..., IT_{|input|}, OT_1) \times \\
    & \pi_{\theta}(OT_3 \;\;|\;\; IT_1, IT_2, ..., IT_{|input|}, OT_1, OT_2) \times \\
    & ... \\
    & \pi_{\theta}(OT_{|output|} \;\;|\;\; IT_1, ..., IT_{|input|}, OT_1, ..., OT_{|output|-1}) \\
    = & \pi_{\theta}(OT_1, OT_2, ..., OT_{|output|} \;\;|\;\; IT_1, IT_2, ..., IT_{|input|}) \\
    = & \pi_{\theta}(MA \;\;|\;\; MS_1)
\end{aligned}
\label{equ:MacroAction1}
\end{equation}
where $IT_i$ and $OT_i$ are input tokens and output tokens, respectively; $MS_1 \triangleq \langle IT_1, IT_2, ..., IT_{|input|} \rangle$ and $MA \triangleq \langle OT_1, OT_2, ..., OT_{|output|} \rangle$ represent the \textbf{macro state} and the \textbf{macro action}, respectively.

In general, the output sequence $\langle OT_1, ..., OT_{|output|} \rangle$ can be partitioned into $K$ segments, yielding generalized macro states and macro actions:
\begin{equation}
\begin{aligned}
    MS_i &\triangleq \langle MS_{i-1}, MA_{i-1} \rangle \\
    MA_i &\triangleq \langle OT_m, OT_{m+1}, ..., OT_{m+n} \rangle
\end{aligned}
\end{equation}
This formulation leads to the following decomposition:
\begin{equation}
\begin{aligned}
    & \pi_{\theta}(MA \;\;|\;\; MS_1) \\
    = & \pi_{\theta}(MA_1 \;\;|\;\; MS_1) \times \\ 
    & \pi_{\theta}(MA_2 \;\;|\;\; MS_1, MA_1) \times \\
    & ... \\
    & \pi_{\theta}(MA_K \;\;|\;\; MS_1, MA_1, MA_2, ..., , MA_{K-1}) \\
   = & \pi_{\theta}(MA_1 \;\;|\;\; MS_1) \times \\ 
    & \pi_{\theta}(MA_2 \;\;|\;\; MS_2) \times \\
    & ... \\
    & \pi_{\theta}(MA_K \;\;|\;\; MS_K) \\ 
   = & \prod_{T=1}^{K} \pi_{\theta}(MA_T \;\;|\;\; MS_T)
\end{aligned}
\label{equ:MacroAction2}
\end{equation}
where $T$ represents the \textbf{macro timestep}.

\subsection{Derivation of the GPG Theorem}
Given the macro states $MS_i$ and macro actions $MA_i$ as defined above, we establish the following Generalized Policy Gradient (GPG) Theorem for Transformer-based policies:
\begin{equation}
\begin{aligned}
    \nabla_{\theta} J(\theta) 
    = & \mathbb{E}_{\tau \sim \pi_\theta}\{ \sum_{T=1}^{K} [\nabla_{\theta}\log\pi_{\theta}(MA_T|MS_T) \Phi_T ] \}
\end{aligned}
\end{equation}

\emph{A principal advantage of the GPG Theorem lies in its accommodation of macro-action segments with arbitrary length}. This flexible formulation yields significant practical benefits: notably, it naturally supports trajectory segmentation using special tokens (e.g., [SEP], [CLS] or [TOOL]). We elaborate on these applications and implementation considerations in Section \ref{sec:PracticalGPG}.

We now present the formal derivation of the Generalized Policy Gradient (GPG) Theorem:
\begin{align}
    & \nabla_{\theta} J(\theta)  \\
    = & \nabla_{\theta} \mathbb{E}_{\tau \sim \pi_\theta}[R(\tau)] \\
    = & \nabla_{\theta} \sum_{\tau}P(\tau;\theta)R(\tau) \\
    = & \sum_{\tau} \nabla_{\theta} P(\tau;\theta)R(\tau) \\
    = & \sum_{\tau} P(\tau;\theta) \frac{\nabla_{\theta} P(\tau;\theta)}{P(\tau;\theta)} R(\tau) \\
    = & \sum_{\tau} P(\tau;\theta) \nabla_{\theta} \log P(\tau;\theta) R(\tau) \\
    = & \sum_{\tau} P(\tau;\theta) \nabla_{\theta} [ \log \mu(s_1) \prod_{t=1}^{H} \pi_{\theta}(a_t|s_t) P(s_{t+1}|s_t,a_t) ] R(\tau) \\
    = & \sum_{\tau} P(\tau;\theta) \nabla_{\theta} [ \log \prod_{t=1}^{H} \pi_{\theta}(a_t|s_t) P(s_{t+1}|s_t,a_t) ] R(\tau) \label{equ:transition1} \\
    = & \sum_{\tau} P(\tau;\theta) \nabla_{\theta} [ \log \prod_{t=1}^{H} \pi_{\theta}(a_t|s_t)] R(\tau) \label{equ:transition2} \\
     = & \sum_{\tau} P(\tau;\theta) \nabla_{\theta} [\log \prod_{T=1}^{K} \pi_{\theta}(MA_T|MS_T) ] R(\tau) \label{equ:transition3}  \\
    = & \sum_{\tau}P(\tau;\theta) [\sum_{T=1}^{K}\nabla_{\theta}\log\pi_{\theta}(MA_T|MS_T) ] R(\tau) \\
    = & \sum_{\tau}P(\tau;\theta) [\sum_{T=1}^{K}\nabla_{\theta}\log\pi_{\theta}(MA_T|MS_T) R(\tau)] \\
    = & \mathbb{E}_{\tau \sim \pi_\theta}\{ \sum_{T=1}^{K} [\nabla_{\theta}\log\pi_{\theta}(MA_T|MS_T) R(\tau)] \} \label{equ:GPG1} \\
    = & \mathbb{E}_{\tau \sim \pi_\theta}\{ \sum_{T=1}^{K} [\nabla_{\theta}\log\pi_{\theta}(MA_T|MS_T) \Phi_T] \} \label{equ:GPG2}
\end{align}

The key steps in the proof are as follows:

1. The equality between Equations (\ref{equ:transition1}) and (\ref{equ:transition2}) follows from the deterministic state transition in Transformer-based policies, where:
\begin{align}
    s_{t+1} = [s_t, a_t] \xrightarrow{} P(s_{t+1}|s_t,a_t) = 1
\end{align}

2. The transition from Equation (\ref{equ:transition2}) to Equation (\ref{equ:transition3}) follows from the autoregressive property of Transformer-based policies, where each state is constructed as $s_{t+1} = [s_t, a_t]$. This leads to the following complete derivation:
\begin{equation}
\begin{aligned}
    & \prod_{t=1}^{H} \pi_{\theta}(a_t|s_t) \\
    = & \pi_{\theta}(a_1|s_1) \times \pi_{\theta}(a_2|s_2) \times ... \times \pi_{\theta}(a_H|s_H)  \\
    = & \pi_{\theta}(a_1|s_1) \times \pi_{\theta}(a_2|s_1, a_1) \times ... \times \pi_{\theta}(a_H|s_0, a_0, a_1, ..., a_{H-1}) \\
    = & \pi_{\theta}(a_1, a_2, ..., a_H|s_1) \\
    = & \pi_{\theta}(MA \;\;|\;\; MS_1) \\
    = & \pi_{\theta}(MA_1 \;\;|\;\; MS_1) \times \\ 
    & \pi_{\theta}(MA_2 \;\;|\;\; MS_1, MA_1) \times \\
    & ... \\
    & \pi_{\theta}(MA_K \;\;|\;\; MS_1, MA_1, MA_2, ..., , MA_{K-1}) \\
    = & \prod_{T=1}^{K} \pi_{\theta}(MA_T \;\;|\;\; MS_T)
\end{aligned}
\end{equation}
% As mentioned in Equation (\ref{equ:MacroAction2}), $T$ represents the macro timestep.

3. The transition from Equation (\ref{equ:GPG1}) to Equation (\ref{equ:GPG2}) mirrors the generalization in standard Policy Gradient Theorem, i.e., from Equation (\ref{equ:PG1}) to Equation (\ref{equ:PG2}).

\subsection{Relation with Existing Methods}
\begin{figure*}[!th]
    \centering
    \includegraphics[width=1.0\linewidth]{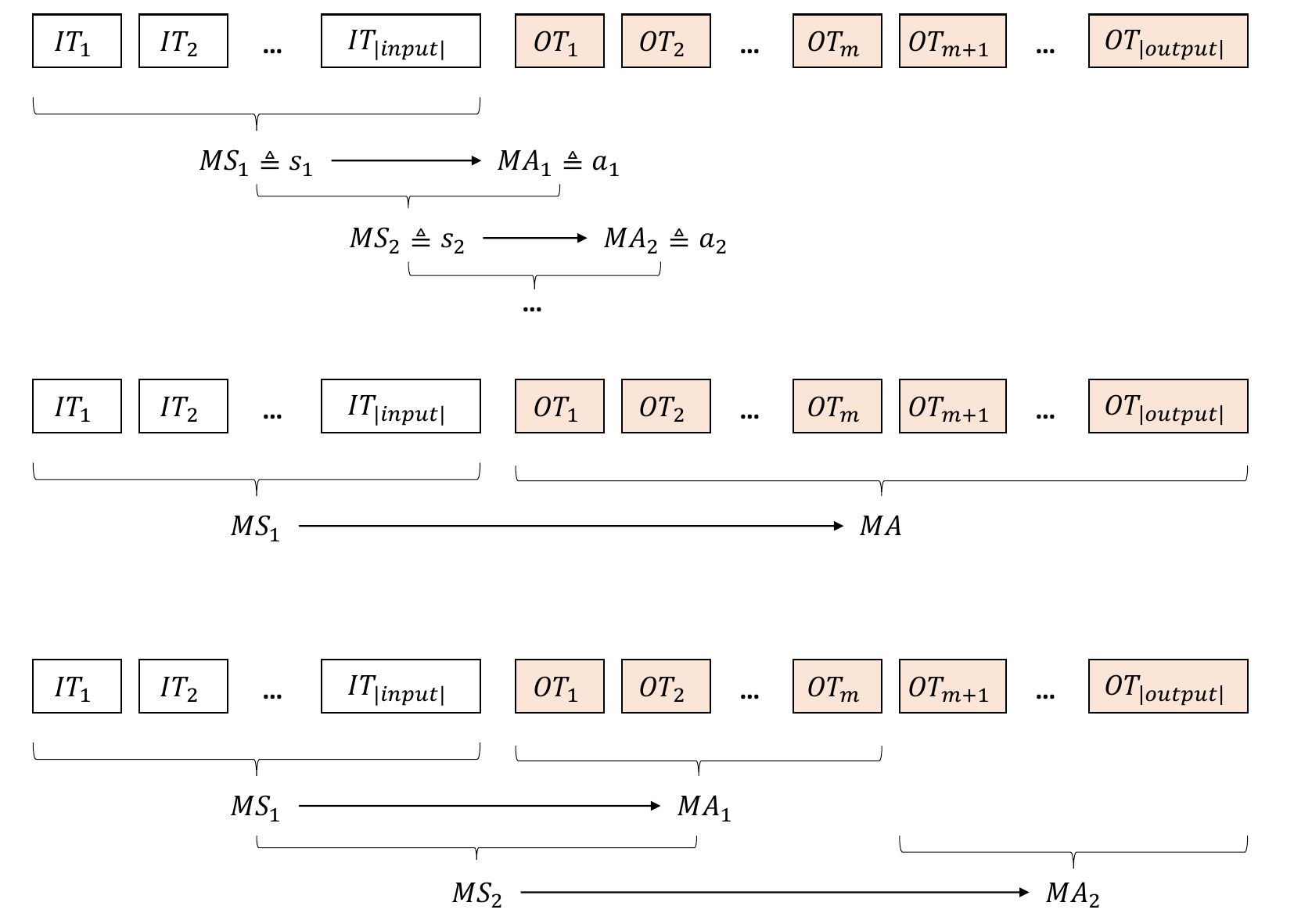}
    \caption{The top shows the special GPG's policy where $K=|output|$, which is exactly the vanilla Transformer-based policy. The middle shows the special GPG's policy where $K=1$, which is similar to the GRPO policy. The bottom shows the general form of GPG, and we take $K=2$ as an example.}
    \label{fig:GPGandOthers}
\end{figure*}

It is straightforward to demonstrate that several existing methods emerge as special cases of our GPG framework, as illustrated in Figure \ref{fig:GPGandOthers}. We identify two particularly important cases:

(1) \textbf{Token-Level Policy Gradient}: When $K=|\text{output}|$ (i.e., each macro-action corresponds to a single output token, $MA_i \triangleq OT_i$), our GPG reduces to:
\begin{equation}
\nabla_\theta J(\theta) = \mathbb{E}\left[\sum_{i=1}^{|\text{output}|} \nabla_\theta \log \pi_\theta(OT_i|MS_i) Q^\pi(MS_i,OT_i)\right] \nonumber
\end{equation}
which is precisely the standard Policy Gradient Theorem, i.e., Equation (\ref{equ:PG2}). This establishes the standard policy gradient as a special case of our generalized framework.

(2) \textbf{Sequence-Level Policy Gradient}: 
When $K=1$ (i.e., the entire output sequence comprises a single macro-action), the framework reduces to: $MA_1  = MA \triangleq \langle OT_1, OT_2, \ldots, OT_{|\text{output}|}\rangle$. This configuration exactly recovers the GRPO paradigm where the complete output sequence functions as an indivisible action unit, reward signals and the the advantages are evaluated over batches of complete output sequences, and gradient steps are performed at the full-sequence abstraction level.

\section{Practical Implementation of GPG Theorem}\label{sec:PracticalGPG}
\begin{figure*}[!t]
    \centering
    \includegraphics[width=0.96\linewidth]{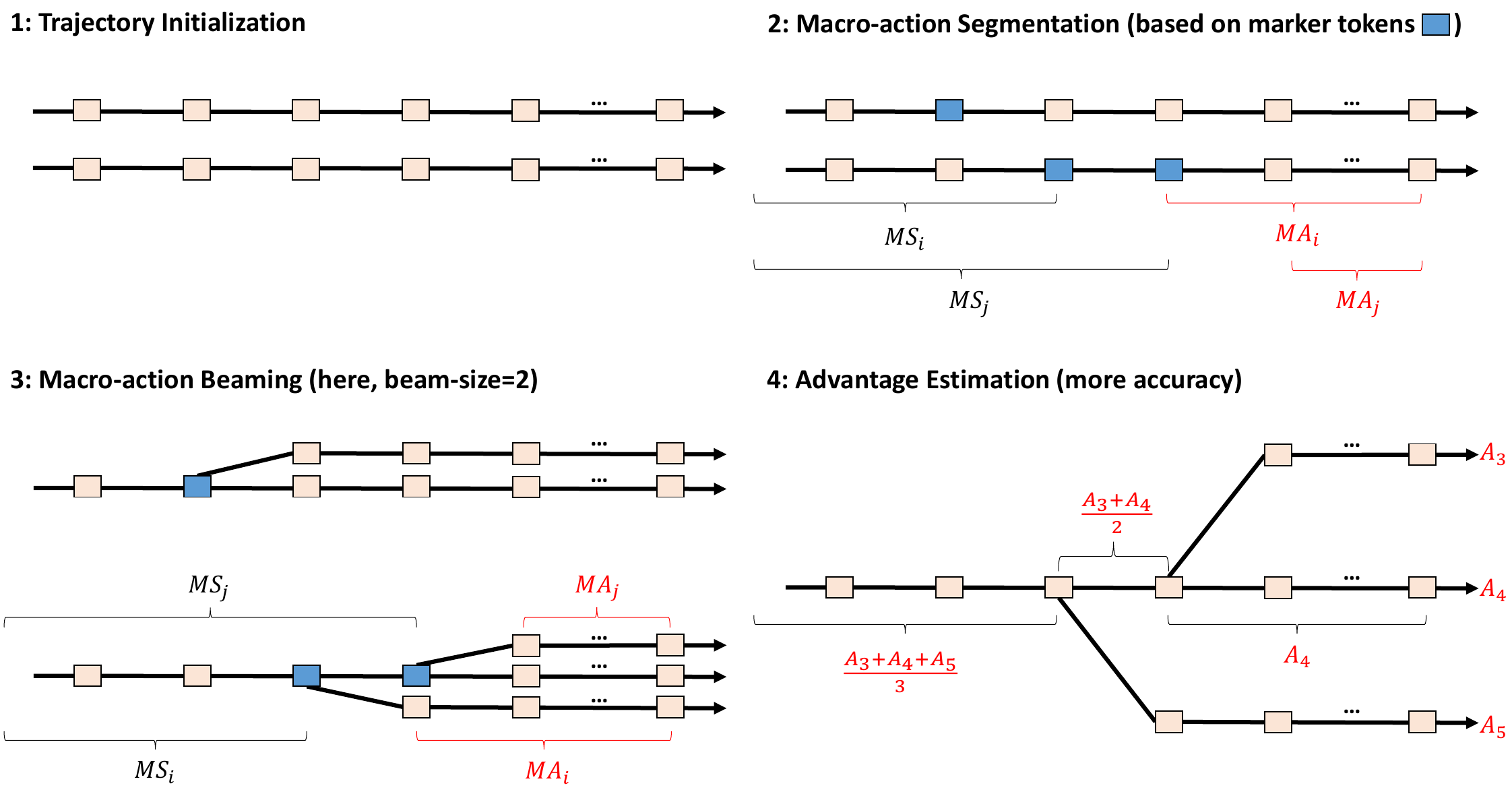}
    \caption{The practical application of the GPG Theorem for LLM-based policy training.}
    \label{fig:GPGpractice}
\end{figure*}

Figure \ref{fig:GPGpractice} illustrates the four-phase pipeline for implementing the GPG Theorem in Transformer-based policy optimization: (1) Trajectory Initialization; (2) Macro-action Segmentation; (3) Macro-action Beaming; (4) Advantage Estimation.

\subsection{Trajectory Initialization}
For a given input query, we initialize multiple trajectories using the policy model $\pi_\theta$, where each trajectory comprises the Transformer's original token-level outputs. This population-based approach enables exploration of the action space.

\subsection{Macro-action Segmentation}
As previously discussed, the GPG Theorem's principal advantage lies in its capacity to accommodate macro-action segmentation of arbitrary length. In practical applications, macro-action segmentation can be performed by identifying \emph{marker tokens}. Below, we present three prototypical use cases, while acknowledging that practitioners may adapt this approach to suit their particular requirements.

\begin{itemize}
  \item \textbf{Agentic Reasoning}: Structured agentic tool-using trajectories containing explicit semantic tags (e.g., $\langle think \rangle$ ... $\langle /think \rangle$, $\langle tool \rangle$ ... $\langle /tool \rangle$) can be segmented at these predefined token boundaries.
  \item \textbf{Document Composition}: In LLM-based text generation, where outputs consist of multiple paragraphs, conventional line breaks (e.g., \texttt{\textbackslash n\textbackslash n}) naturally serve as effective segmentation points between compositional units.
  \item \textbf{Creative Problem-Solving}: Unlike routine tasks, creative processes often present greater challenges for LLMs. Here, high-entropy tokens (those exhibiting greater predictive uncertainty) may be employed as segmentation boundaries. 
\end{itemize}

\subsection{Macro-action Beaming}
For each marker token at position $t$:
\begin{itemize}
\item Macro-state is the partial trajectory before this token, i.e., $MS = \langle IT_1, IT_2, ..., IT_{|input|}, OT_1, ..., OT_t \rangle$
\item Macro-action is the partial trajectory after this token, i.e., $MA = \langle OT_{t+1}, ..., OT_{|output|} \rangle$
\end{itemize}
In the macro-action beaming step, we further generate $N$ candidate continuations ($MA^{(1)}, ..., MA^{(N)}$) from each $MS$, enabling diverse exploration.

\subsection{Advantage Estimation}
We propose calibrated advantage computation, which improves upon GRPO's approach:
\begin{itemize}
  \item \textbf{Reward Computation}: Obtain trajectory rewards $r_i$ via reward model or rule-based scoring.
  \item \textbf{Initial Advantage}: $A_i^{init} = \frac{r_i - \mu_{\mathcal{G}}}{\sigma_{\mathcal{G}}}$ where $\mathcal{G}$ is the trajectory group.
  \item \textbf{Token-level Calibration}: For token $a_t$ in trajectory $i$:
    \begin{equation}
        A_t = \frac{1}{|\mathcal{S}t|} \sum_{j\in\mathcal{S}_t} A_j^{init}
    \end{equation}
    where $\mathcal{S}t$ is the set of trajectories sharing prefix $a_{[1:t-1]}$
\end{itemize}
% In GRPO, there is no advantage calibration: all Transformer's output tokens in the same trajectory share the same advantage. However, it is easy to see that different macro-actions sharing this same macro-state may result in different rewards and thus advantages. Our calibration can get a better advantage estimation for this situation.

\subsection{Policy Optimization}
Using the calibrated advantages, we can use the classical RL methods like PPO to optimize the Transformer-based policy's model parameters as shown by Equation (\ref{equ:PPO}).

\section{Experiment}
We validate our GPG framework on agentic reasoning tasks - specifically, LLM-based tool-use agent training - selected for two principal advantages: (1) Predefined semantic tags provide deterministic segmentation boundaries, removing the ambiguity of learned marker tokens; (2) The complex action space and delayed rewards in tool manipulation present a rigorous testbed for policy optimization methods. We term our approach Agentic Reinforced Policy Optimization (ARPO) to reflect its specialized application domain \footnote{See more details in our paper: \url{https://arxiv.org/abs/2507.19849}}.

\subsection{Dataset}
We evaluate our approach on two challenging categories of long-horizon agentic reasoning tasks. 

\textbf{Mathematical Reasoning:} GSM8K, MATH \citep{MATH}, MATH500 \citep{math500}, AIME2024 and AIME2025 \footnote{\url{https://huggingface.co/datasets/AI-MO/aimo-validation-aime}}. 

\textbf{Knowledge-Intensive Reasoning:} WebWalker \citep{2501_WebWalker}, HotpotQA \citep{hotpotqa}, 2WikiMultihopQA \citep{2wiki}, Musique \citep{musique} and bamboogle~\citep{bamboogle}.

\subsection{Baselines.} To comprehensively assess the performance of ARPO, we employ the following three baseline methodologies.

\textbf{Direct Reasoning Approaches:} We evaluate instruction-tuned versions of both the Qwen2.5~\citep{qwen2.5} and Llama3.1~\citep{llama3} model families.

\textbf{Trajectory-level RL Methods:} We conduct comparative evaluations between ARPO and established trajectory-level RL algorithms commonly employed for training LLM-based tool-use agents, including GRPO~\citep{deepseekmath}, DAPO~\citep{yuDAPO}, and the REINFORCE++ ~\citep{hu2025reinforce++}.

\subsection{Evaluation Protocol} 
To ensure consistency with established reasoning benchmarks, we employ a browser-enabled search engine and a python code interpreter as tools for evaluation. For knowledge-intensive reasoning tasks, we measure accuracy using F1 scores. All other tasks are assessed under the LLM-as-Judge paradigm using the Qwen2.5-72B-instruct model.

We implement pass@1 evaluation with stochastic sampling, configuring the temperature parameter to 0.6 and top-p to 0.95, respectively. Following prior methodology~\citep{searcho1}, we extract model responses by identifying text segments delimited by \texttt{\textbackslash box{}} markers in the output. This standardized approach ensures fair and reproducible comparison across all evaluated tasks.

\begin{table*}[t]
\centering
\resizebox{1.0\columnwidth}{!}{
\begin{tabular}{p{2.8cm}ccccccccccc}

\hline
\multirow{2}[2]{*}{\textbf{Method}} & \multicolumn{5}{c}{\textbf{Mathematical Reasoning}} & \multicolumn{5}{c}{\textbf{Knowledge-Intensive Reasoning}} & \multirow{2}[2]{*}{\textbf{Avg.}} \\
& AIME24 & AIME25 & MATH500 & GSM8K & MATH & WebWalker & HQA & 2Wiki. & MuSiQ. & Bamb. & \\

\hline
\textbf{Qwen2.5-3B} & 10.0 & 6.7 & 63.0 & 75.0 & 71.6 & 0.5 & 9.7 & 9.4 & 3.6 & 11.7 & 26.1 \\
\quad + TIR Prompting                 & 6.7 & 6.7 & 52.2 & 56.6 & 62.8 & 14.0 & 15.4 & 14.1 & 6.1 & 16.4 & 25.1 \\
\quad + GRPO                 & \underline{20.0} & 13.3 & \textbf{72.0} & \textbf{86.0} & 81.0 & \underline{21.0} & \underline{56.5} & \underline{64.5} & 24.7 & 65.2 & 50.4 \\
\quad + Reinforce ++         & 16.7 & 13.3 & 70.4 & \underline{85.0} & 80.2 & 19.5 & 55.9 & 62.3 & 27.9 & \underline{65.7} & 49.7 \\
\quad + DAPO                 & \underline{20.0} & \underline{16.7} & 71.2 & \underline{85.0} & \underline{81.2} & 19.5 & 54.8 & 62.5 & \textbf{30.0} & 64.8 & \underline{50.6} \\
\rowcolor[RGB]{236,244,252} 
\quad + ARPO                 & \textbf{23.3} & \textbf{20.0} & \underline{71.4} & \underline{85.0} & \textbf{82.5} & \textbf{24.5} & \textbf{58.5} & \textbf{67.4} & \underline{28.7} & \textbf{66.8} & \textbf{52.8} \\

\hline
\textbf{Qwen2.5-7B} & 10.0 & 10.0 & 70.6 & 90.2 & 82.0 & 2.0 & 12.2 & 12.6 & 6.6 & 24.0 & 32.0 \\
\quad + TIR Prompting        & 6.7 & 10.0 & 68.2 & 64.6 & 78.2 & 15.5 & 14.8 & 18.3 & 9.5 & 23.6 & 31.0 \\
\quad + GRPO                 & 23.3 & \underline{26.7} & 78.0 & \textbf{92.8} & \underline{87.8} & 22.0 & \textbf{59.0} & \textbf{76.1} & \underline{30.6} & \underline{68.4} & \underline{56.5} \\
\quad + Reinforce ++         & \underline{26.7} & 23.3 & 78.0 & \underline{92.2} & \textbf{88.8} & \textbf{26.0} & 55.1 & \underline{68.9} & 25.2 & 64.9 & 54.9 \\
\quad + DAPO                 & 20.0 & 23.3 & \textbf{80.4} & 91.0 & \textbf{88.8} & \underline{24.0} & 57.7 & 68.4 & 28.6 & 65.5 & 54.8 \\
\rowcolor[RGB]{236,244,252} 
\quad + ARPO                 & \textbf{30.0} & \textbf{30.0} & \underline{78.8} & \underline{92.2} & \textbf{88.8} & \textbf{26.0} & \underline{58.8} & \textbf{76.1} & \textbf{31.1} & \textbf{71.5} & \textbf{58.3} \\

\hline
\textbf{Llama3.1-8B} & 3.3 & 0.0 & 43.3 & 81.4 & 60.6 & 3.0 & 24.3 & 24.6 & 10.4 & 40.0 & 28.8 \\
\quad + TIR Prompting        & 3.3 & 3.3 & 39.4 & 73.8 & 58.2 & 15.0 & 48.5 & 47.5 & 15.5 & 58.4 & 36.3 \\
\quad + GRPO                 & 13.3 & \underline{13.3} & \underline{62.4} & \underline{87.4} & \underline{79.2} & 26.5 & \underline{57.8} & \underline{71.8} & \underline{31.0} & 68.2 & \underline{51.1} \\
\quad + Reinforce ++         & 13.3 & \textbf{16.7} & 61.4 & 87.0 & 77.2 & \underline{27.5} & 57.1 & 71.6 & 29.9 & \underline{69.1} & \underline{51.1} \\
\quad + DAPO                 & \underline{16.7} & \underline{13.3} & 61.2 & \underline{87.4} & 76.4 & 25.5 & 56.6 & 70.3 & 29.2 & 67.3 & 50.4 \\
\rowcolor[RGB]{236,244,252} 
\quad + ARPO                 & \textbf{23.3} & \textbf{16.7} & \textbf{64.6} & \textbf{88.0} & \textbf{80.2} & \textbf{30.5} & \textbf{65.4} & \textbf{75.5} & \textbf{34.8} & \textbf{73.8} & \textbf{55.3} \\

\hline
\end{tabular}
}
\caption{The comprehensive evaluation results across 10 challenging reasoning tasks are presented. For clarity, we highlight the top two performing methods using \textbf{bold} and \underline{underline} formatting. The following dataset abbreviations are used: HQA (HotpotQA), 2Wiki. (2wikiMultiHopQA), MuSi. (MuSiQue), and Bamb (Bamboogle).}
\label{tab:main_table}
\end{table*}

\subsection{Main Results}
The key experimental results are presented in Table~\ref{tab:main_table}. Under standardized evaluation conditions, ARPO demonstrates consistent superiority over all trajectory-level RL baselines, with the following critical observations.

\textbf{Inherent Limitations of Prompting Approaches:} Our analysis reveals that Tool-integrated prompting (TIR) methods~\citep{searcho1} exhibit fundamental constraints in discovering optimal tool-use strategies. Across both Qwen and Llama model families, TIR prompts yield marginal performance gains that frequently underperform direct reasoning baselines. These results indicate that prompt engineering alone cannot effectively guide LLMs toward sophisticated tool manipulation while preserving their native reasoning abilities.

\textbf{Challenges in Trajectory-Level Optimization:} The comparative analysis exposes significant limitations in conventional trajectory-level RL algorithms. While DAPO shows competence in single-turn reasoning, its performance degrades markedly in multi-turn tool interaction scenarios, particularly for knowledge-intensive tasks. This empirical evidence corroborates our hypothesis that trajectory-level approaches are fundamentally constrained in facilitating granular, step-wise tool behavior learning in LLMs.

\textbf{Consistent Superiority of ARPO:} ARPO achieves state-of-the-art performance across all 10 benchmark datasets, delivering an average accuracy gain of 4\% over competing methods while maintaining robust performance across diverse domains. Notably, the framework demonstrates remarkable backbone-agnostic properties, showing substantial improvements for both Qwen and Llama model series. These results validate ARPO's effectiveness as a versatile and adaptable solution for tool-augmented language models.

\textbf{Model Capacity Scaling:} The results demonstrate consistent performance improvements as model size increases across all methods. Notably, ARPO achieves the most significant gains from scaling, with 7B models showing 5.5\% absolute improvement over 3B variants (58.3\% vs 52.8\%). The performance delta between ARPO and baselines widens with larger models (e.g., +7.8\% over DAPO for 7B vs +2.2\% for 3B). Llama3.1-8B shows particular sensitivity to optimization methods, with ARPO delivering 4.2\% improvement over the next best method.

In summary, these results collectively demonstrate that ARPO's step-level optimization paradigm fundamentally addresses key limitations in existing approaches for tool-augmented language models. The consistent performance advantages across model sizes and task categories suggest the framework's robustness and generalizability.

%\subsection{Case Study}

\section{Related Work}
\subsection{Policy Gradient in Reinforcement Learning} 
The Policy Gradient Theorem \cite{sutton1999policy} provides the foundational framework for gradient-based policy optimization in reinforcement learning. The simplest form, REINFORCE \cite{williams1992simple}, estimates the gradient as $\nabla_\theta J(\theta) = \mathbb{E}[\nabla_\theta \log \pi_\theta(a|s) Q^\pi(s,a)]$, where $\pi_\theta$ represents the policy parameterized by $\theta$, and $Q^\pi(s,a)$ is the long-term value of taking action $a$ at state $s$. While straightforward, this approach suffers from high variance and inefficient exploration. To address these limitations, several advanced variants have been developed. Natural Policy Gradient \cite{kakade2001natural} incorporates the Fisher information matrix $F(\theta)$ to enable invariant updates: $\tilde{\nabla}\theta J(\theta) = F(\theta)^{-1} \nabla\theta J(\theta)$. This provides more stable convergence by accounting for the underlying geometry of the parameter space. Deterministic Policy Gradient (DPG) \cite{silver2014deterministic} extends the framework to deterministic policies $\mu_\theta: \mathcal{S} \rightarrow \mathcal{A}$, with gradient $\nabla_\theta J(\theta) = \mathbb{E}[\nabla_\theta \mu_\theta(s) \nabla_a Q^\mu(s,a)|_{a=\mu_\theta(s)}]$. This is particularly effective in continuous action spaces. Trust Region Policy Optimization (TRPO) \cite{schulman2015trust} constrains policy updates within a trust region to guarantee monotonic improvement, while Proximal Policy Optimization (PPO) \cite{schulman2017proximal} simplifies this through clipped objective functions.

\subsection{Policy Gradient in LLM Optimization} 
Recent advances have significantly advanced policy optimization techniques for language tasks, with Reinforcement Learning from Human Feedback (RLHF) and Reinforcement Learning with Verifiable Rewards (RLVR)~\citep{rlhf,tulu3} emerging as dominant paradigms for large language model alignment. These approaches have demonstrated remarkable success in aligning model outputs with human preferences while maintaining generation quality. Building upon these foundations, Generalized Reinforcement Policy Optimization (GRPO) introduces enhanced advantage estimation through grouped trajectory analysis within the PPO framework, offering improved stability in complex language generation tasks. Direct Preference Optimization (DPO) \cite{rafailov2023direct} bypasses explicit reward modeling by optimizing policies directly from offline preference datasets, while Decoupled Clip and Dynamic sAmpling Policy Optimization (DAPO) \cite{yuDAPO} exploring algorithm design across various RL modules. 

\subsection{Agentic Policy Optimization}
There are different categories of agentic policy optimization. Some focus on prompt optimization for LLM-based agents, e.g., TPTU \cite{ruan2023tptu}; some for supervised training LLMs for agentic tool-using, e.g., TPTU-v2 \cite{kong2024tptu}; some for cooperation within multiple LLM-based agents, e.g., LLaMAC \cite{zhang2023controlling}; some for structured data like sheet and sql operation \cite{zhang2024benchmarking,li2024pet,yang2024sql,chen2025sheetagent,wang2025general}; some for unstructured data retrieval like agentic rag \cite{li2024dmqr,cheng2025dualrag}. Recently, using RL to optimize LLMs for agentic tool-using is very popular, e.g., tool-star \cite{dong2025toolstar}, arpo \cite{dong2025arpo}, and aepo \cite{dong2025aepo}; the work in this paper presents a foundation theory for this kind of agentic policy optimization.

\subsection{Our Contribution}
The existing methodological advances have substantially pushed the boundaries of policy optimization, especially in complex environments characterized by high-dimensional state and action spaces. However, existing approaches remain generic solutions applicable to all policy types, failing to specifically account for the unique auto-regressive nature of Transformer architectures that underpin modern LLMs. This represents a critical limitation, as the sequential decision-making process in Transformers differs fundamentally from conventional reinforcement learning settings.

Our work bridges this gap through two key contributions: First, the proposed GPG Theorem is specifically designed for Transformer-based policies, explicitly modeling their auto-regressive properties. Second, we develop ARPO as an instantiation of the GPG Theorem, providing the first policy optimization framework that natively respects the architectural constraints of LLMs while being specifically tailored for agentic reasoning tasks. This represents a paradigm shift from previous general-purpose policy optimization methods to architecture-aware optimization for LLMs.

\section{Conclusion}
This paper presents the Generalized Policy Gradient (GPG) Theorem, a novel theoretical framework specifically tailored for optimizing Transformer-based policies. Our analysis establishes that both the conventional Policy Gradient Theorem and GRPO can be naturally derived as special cases within our unified GPG framework. Beyond theoretical contributions, we provide practical implementation guidelines for effectively applying GPG to LLM training scenarios. % Comprehensive evaluations on LLM-based agentic reasoning tasks demonstrate the superior performance of our approach compared to GRPO and other policy gradient baselines. These findings advance our understanding of efficient policy optimization techniques for modern language models.
Comprehensive evaluations demonstrate the superior performance of our approach. These findings advance our understanding of efficient policy optimization techniques for modern language models.

\bibliographystyle{plain}
\bibliography{allcite}

\end{document}